\documentclass{article}
\usepackage{spconf,amsmath,epsfig}
 
\usepackage{times}  

\usepackage[hyphens]{url}  
\usepackage{graphicx} 

\usepackage{caption} 
\usepackage{algorithm}
\usepackage{algorithmic}
\usepackage{makecell}
\usepackage{wrapfig}
\usepackage{vcell}
\usepackage{array}
\usepackage[switch]{lineno}
\usepackage{times}
\usepackage{epsfig}
\usepackage{amsmath}
\usepackage{amssymb}
\usepackage{multirow}
\usepackage{bbding}
\usepackage{microtype}      
\usepackage{xcolor}         
\usepackage[justification=centering]{subcaption}
\usepackage{exscale}
\usepackage{hyperref}
\hypersetup{hypertex=true,
            colorlinks=true,
            linkcolor=red,
            anchorcolor=red,
            citecolor=black}

\usepackage{relsize}
\usepackage{xspace}

\makeatletter
\DeclareRobustCommand\onedot{\futurelet\@let@token\@onedot}
\def\@onedot{\ifx\@let@token.\else.\null\fi\xspace}

\def\eg{\emph{e.g}\onedot} 
\def\ie{\emph{i.e}\onedot} 
 
\def\etc{\emph{etc}\onedot}

\newcommand{\saveCurrentHref}[1]{%
  \global\expandafter\let
    \csname saved@Href@#1\endcsname\Hy@footnote@currentHref
  \ignorespaces
}
\newcommand{\restoreCurrentHref}[1]{%
  \expandafter\let
    \expandafter\Hy@footnote@currentHref\csname saved@Href@#1\endcsname
  \ignorespaces
}

\makeatother

\DeclareMathOperator*{\argmin}{argmin}
\newcolumntype{C}{>{\fontfamily{fdr}\fontseries{b}\selectfont}c}
\newcolumntype{T}{>{\fontfamily{fdm}\selectfont\small}c}
\newcolumntype{P}{>{\fontfamily{fdm}\selectfont\small}p{3cm}}

\newcommand{\norm}[1]{\left\lVert#1\right\rVert}

\let\OLDthebibliography\thebibliography
\renewcommand\thebibliography[1]{
  \OLDthebibliography{#1}
  \setlength{\parskip}{0pt}
  \setlength{\itemsep}{0pt plus 0.3ex}
}

\begin{document}\sloppy


\def\x{{\mathbf x}}
\def\L{{\cal L}}

\title{Self-Supervised Face Image Restoration with a One-Shot  Reference}


\twoauthors
 {Yanhui Guo, Fangzhou Luo}
{McMaster University, Canada\\
 \{guoy143,luof1\}@mcmaster.ca}
 {Shaoyuan Xu}
	{Amazon, United States\\
 shaoyux@amazon.com}

\maketitle

%
\begin{abstract}
For image restoration, methods leveraging priors from generative models have been proposed and demonstrated a promising capacity to robustly restore photorealistic and high-quality results. However, these methods are susceptible to semantic ambiguity, particularly with images that have obviously correct semantics such as facial images. In this paper, we propose a semantic-aware latent space exploration method for image restoration (SAIR). By explicitly modeling semantics information from a given reference image, SAIR is able to reliably restore severely degraded images not only to high-resolution and highly realistic looks but also to correct semantics. Quantitative and qualitative experiments collectively demonstrate the superior performance of the proposed SAIR. Our code is available in \url{https://github.com/Liamkuo/SAIR}.

\end{abstract}
\section{Introduction}
Recently, CNN-based \cite{zhang2018image} and Transformer-based \cite{liang2021swinir} image restoration methods have rapidly become the dominant choice for almost all image restoration tasks, such as superresolution, denoising, deblurring and \etc. 
These models are trained by a sample-wise differentiable distortion loss function, such
as $\ell_2$-norm and $\ell_1$-norm metrics, possibly with some regularization term(s). 
However, this straightforward supervised learning usually gives rise to the well-known generalization problem, where the performance decreases dramatically if the assumed degradation models in training mismatch those of the images at the inference stage \cite{luo2021functional,guo2022tip}.

To develop more robust image restoration technologies, generative prior based methods have been proposed recently, including the GAN prior \cite{menon2020pulse,richardson2020encoding,Yang2021GPEN} and Diffusion prior \cite{saharia2022image,wang2023dr2}. 
The GAN prior based methods exploit the pre-trained GAN models, such as BigGAN and StyleGAN \cite{Karras2020ada}, constructing a manifold of high quality images. Following the GAN framework, they estimate a latent vector on the manifold from an input low-quality (LQ) image and the obtained latent vector can be mapped to a high-quality (HQ) image by a pretrained generator \cite{Karras2020ada}. 
For the Diffusion prior based methods \cite{saharia2022image,wang2023dr2}, these technologies diffuse the input images to a  Gaussian noise status 
and restore the HQ images through iterative denoising steps \cite{wang2023dr2}. These generative prior based methods can produce visually good results, even for extremely degraded images \eg $\times$32 downsampling. 
However, these generative prior based methods have a common and fatal problem in that they could cause semantic confusion due to the lack of explicitly modeling the semantic information of the LQ images. For instance, in the case of face image restoration, if the face in the restored image cannot be recognized, then it is a failure of restoration, regardless of how high a visual quality this kind of method can achieve. 
Such a limitation greatly impedes the usefulness of the generative prior based methods for image restoration. 

In this paper, to address the above valuable yet challenging problem, 
we propose a semantic-aware image restoration method (SAIR) to restore severely degraded LQ images $Y$ in a self-supervised manner. 
To reinstall the correct semantics, it is nontrivial to estimate a semantic-consistent latent representation $w$ from $Y$ to reconstruct the HQ images $X$ against $Y$. 
From the perspective of information bottleneck \cite{IB2000}, 
an ideal latent representation $w$ of $Y$ should preserve the minimum necessary 
information to restore an image close to the HQ image $X$ but eliminate the redundant information (\eg noise) of the LQ observation $Y$. 
This indicates an optimization principle that aims to maximize the mutual information $I(w;X)$ while retaining $I(w;Y)$ small. 
However, in a self-supervised framework, the HQ version $X$ is not available. To 
tackle this intractable problem, we can maximize the bottom boundary of $I(w;X)$ by considering the semantic information $\mathrm{SI}(Y)$ from the available LQ observation $Y$, where $\mathrm{SI}$ denotes the semantic information. In this sense, finding the $w$ can be achieved by maximizing $I(w;\mathrm{SI}(Y))$, where $I(w;\mathrm{SI}(Y)) \leq I(w;\mathrm{SI}(X)) \leq I(w;X)$. 
But if the observation $Y$ is severely degraded, it leads to $I(w;\mathrm{SI}(Y)) \ll I(w;\mathrm{SI}(X))$ owing to the loss of semantic information so that the reconstructed HQ images could 
suffer semantic confusion. 
To address this problem, we resort to the additional information $R$ provided by a chosen reference. 
Assuming that abundant semantic information from the reference can be given, the mutual information satisfies $I(w;\mathrm{SI}(R,Y)) \textgreater I(w;\mathrm{SI}(Y))$ and $I(w;\mathrm{SI}(R,Y))\approx I(w;\mathrm{SI}(X))$. 
Hence, we can maximize $I(w;\mathrm{SI}(R,Y))$ to find a good latent representation to restore an HQ image with correct semantics that are shared by the reference and the unknown ground truth $X$. 

\begin{figure}[h]
\vspace{-10pt}
  \centering
  \includegraphics[width=0.4\textwidth]{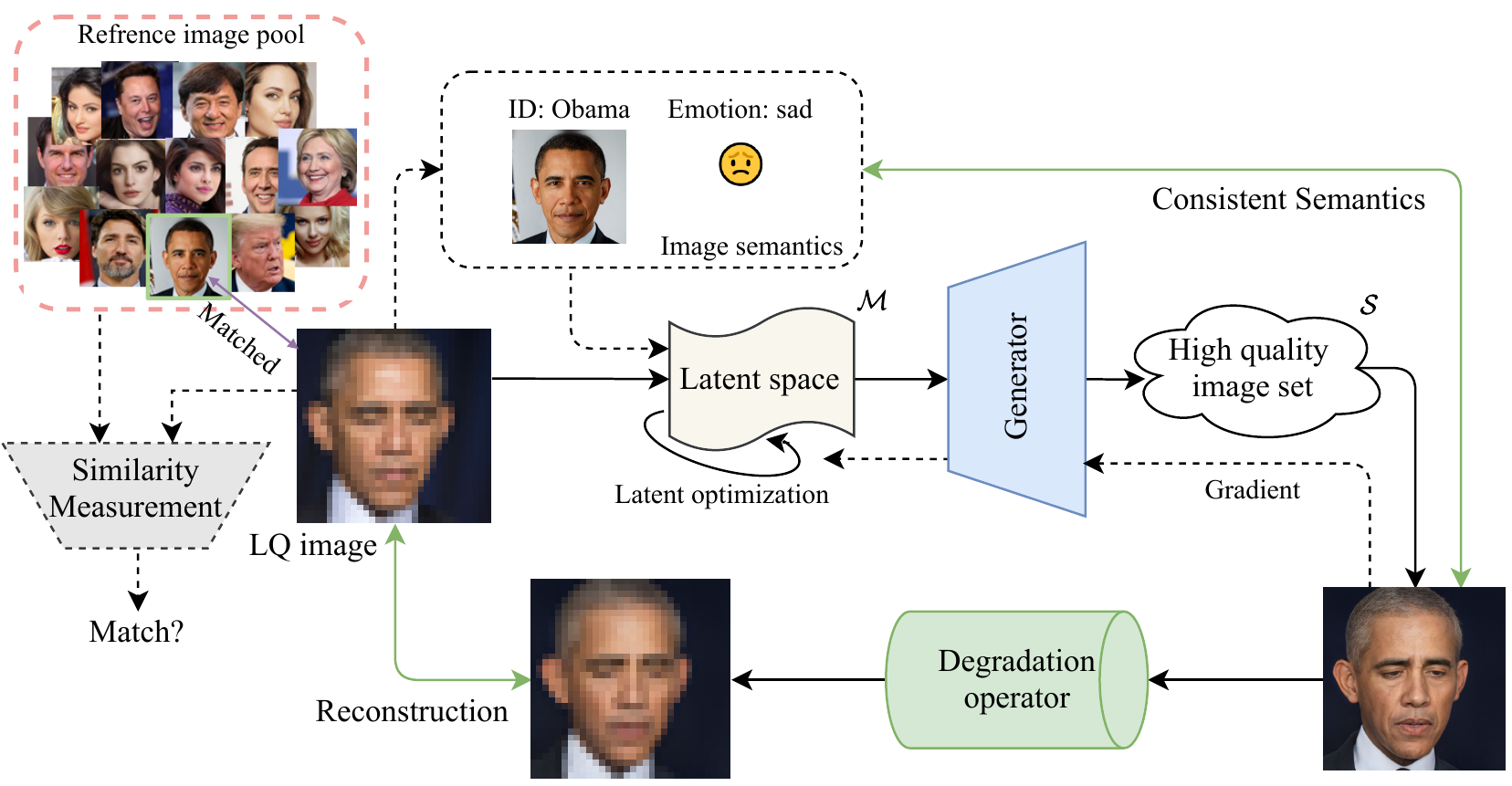}
  \vspace{-10pt}
  \caption{The application of the proposed SAIR method for face restoration.}
  \label{SAIRframework}
\vspace{-0.8cm}
\end{figure}

In this work, we focus on the application of SAIR to the face restoration problem. As depicted in Fig.~\ref{SAIRframework}, an unpaired reference face image with the same identity as the LQ image, more often than not, is available. 
This reference image can be picked out from a candidate image pool by matching the facial similarity provided by the user or a face recognition network. 
In addition, for the face photo, the emotion of the LQ image can be inferred from the rough facial details so that the suppositional emotion can be 
exploited to draw the restored facial details nearer to those of the true facial image. 
Intuitively, the more provided semantic reference information $R$, the more correct semantic details can be achieved. Thus, we take into account the accessible semantics including identity, emotion, and skin color. 
In SAIR, we first build an image manifold $\mathcal{M}$ (a.k.a latent space \cite{Abdalcvpr2020}) for a set of high quality images $\mathcal{S}$. By traversing the manifold guided by the reference information $R$, SAIR can consistently find solutions to generate high quality images 
that have correct semantics without any supervised training. 
Although we focus on face image restoration, our technology is scalable to other types of images. 
For example, the image manifold can be specified to cat images, and the semantic reference $R$ could be fur color, cat breed, and \etc.

\vspace{-10pt}
\section{Approach}
Given the reference $R$ and the LQ observation $Y$, we find the restored HQ image $\hat{X}$ by optimizing the following objective 
\vspace{-5pt}
\begin{equation}
  \hat{X} = \argmin_{X} \mathcal{L}(Y,D(X))  + \mathcal{L}_{\mathrm{SI}}(\mathrm{SI}(R,Y),\mathrm{SI}(X))
  \label{eq_minmum}
\vspace{-7pt}
\end{equation}
which consists of the degradation fidelity term $\mathcal{L}(Y,D(X))$ with a suitable differential metric $\mathcal{L}(\cdot,\cdot)$, 
and the semantic-aware regulation term $\mathcal{L}_{\mathrm{SI}}(\mathrm{SI}(R,Y),\mathrm{SI}(X))$, where $\mathcal{L}_{\mathrm{SI}}(\cdot,\cdot)$ denotes  
a measurement for the semantic difference. 

To utilize the powerful generative model prior, rather than directly solve for the latent image in the pixel space, we optimize a latent code in a GAN face manifold $\mathcal{M}$, parameterized by  $\mathcal{W}^{+} \subseteq \mathbb{R}^{18\times512}$ \cite{Abdal2019iccv}, which is a concatenation of 18 different 512-dimensional
vectors. In this latent space, a random vector $w \in \mathcal{W}^{+}$ can be mapped to the high quality face image using 
a pre-trained generator network $G(w)$ \cite{Karras2020ada}. 
\begin{figure}
  \centering
  \includegraphics[width=0.42\textwidth]{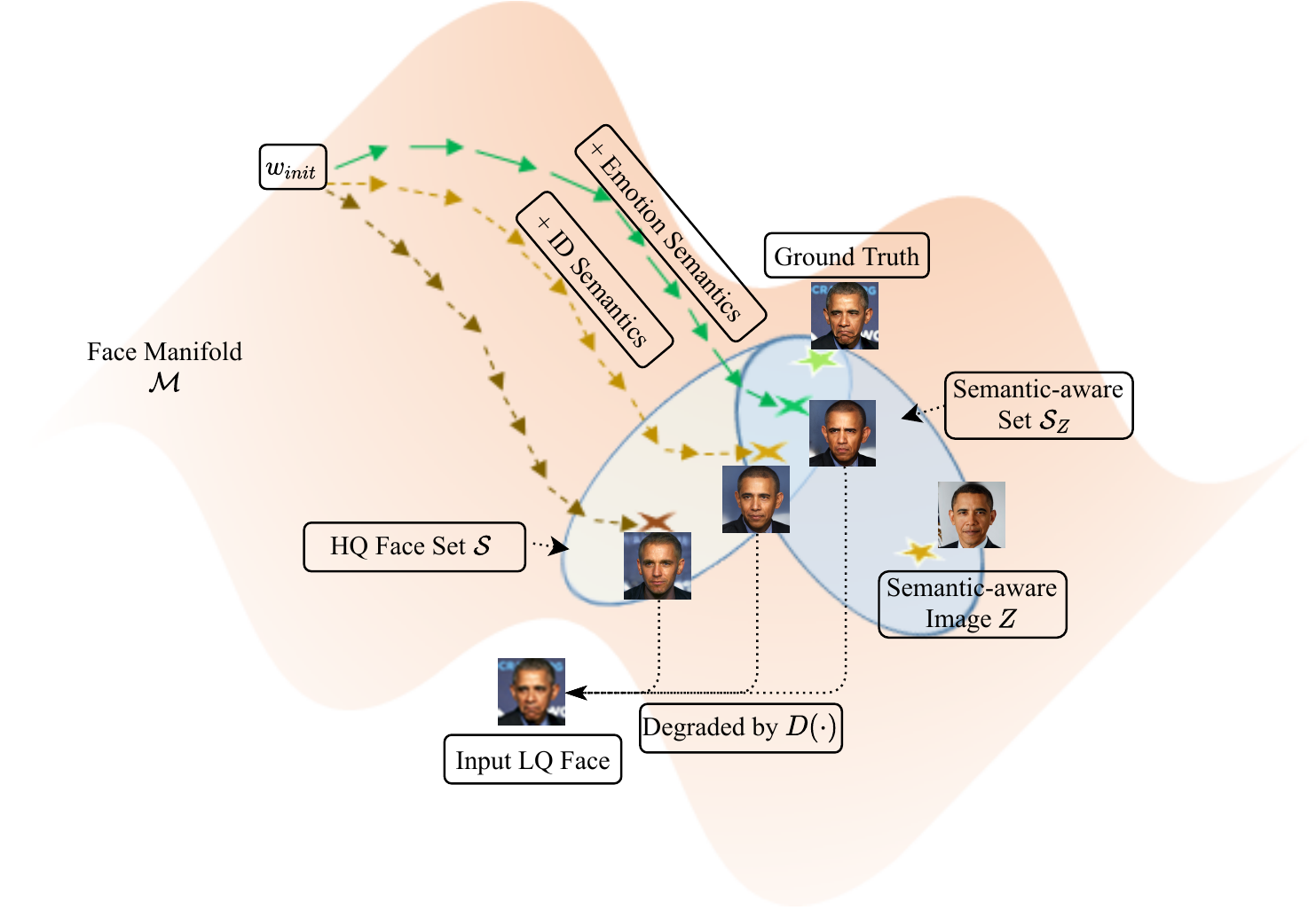}
  \vspace{-10pt}
  \caption{Illustration of the optimization process on the face manifold w/ and w/o the semantic-aware information.}
  \label{fig_intro}
\vspace{-0.5cm}
\end{figure}
With the GAN prior, the optimization variable of SAIR is the latent code $w$ rather than $X$. 
Eq.~(\ref{eq_minmum}) can be restated as follows
 \vspace{-5pt}
\begin{equation}
  \min_{w\in\mathcal{W}^{+}} \mathcal{L}(Y,D(G(w)))  + \mathcal{L}_{\mathrm{SI}}(\mathrm{SI}(R,Y),\mathrm{SI}(G(w)))
  \label{eq_minmum_w}
\vspace{-6pt}
\end{equation}
where $\mathcal{L}_{\mathrm{SI}}(\mathrm{SI}(R,Y),\mathrm{SI}(G(w)))$ consists of three kinds of semantic information \ie the face's identity, the facial expression, and facial skin color. 
To model the abstract semantics, we take account of a reference image $Z$ that has the same identity as $Y$, and an emotion one-hot vector. 
Fig.\ref{fig_intro} illustrates the semantic-aware optimization process of SAIR. 
If without any reference information, the estimated HQ face $\hat{X}$ is located in the area of HQ face set $\mathcal{S}$. 
The set $\mathcal{S}$ contains an infinite number of solutions; it can be greatly squeezed by the given reference image $Z$ to the region $\mathcal{S} \cap \mathcal{S}_Z$, where the set $\mathcal{S}_Z$ is determined by the reference image $Z$.
Adding the facial emotion semantics can further specify the facial details in the restored image and force the final result closer to the true solution. 

\noindent\textbf{Degradation fidelity}
we defines the degradation fidelity term in Eq~(\ref{eq_minmum_w}) as follows  
\vspace{-10pt}
\begin{equation}
  \resizebox{.85\hsize}{!}{$\mathcal{L}_d(Y,w) = \left\|Y - D(\hat{X})\right\|^2_F = \left\|Y - D(G(w))\right\|^2_F$}
  \label{d_loss_1}
  \vspace{-7pt}
\end{equation}
where the degradation operator $D(\cdot)$ of a real-world LQ image can usually be approximately modeled by 
\vspace{-4pt}
\begin{equation}
  D(\hat{X}) = ((\hat{X}\otimes \rm{k})\downarrow_s + \rm{n}_{\sigma})_{JPEG_q}
  \label{degradation_func}
  \vspace{-4pt}
\end{equation}
where $\rm{k}$ and $\rm{n}_{\sigma}$ represent the blur kernel and the noise. $\otimes$, $\downarrow_s$ and $\rm{JPEG_q}$ are the convolution operator, the downsampler with a factor $s$ and the JPEG compression operator with a quality factor $q$, respectively. 
This simple model cannot accurately mimic all complex degradations, 
but even using it, SAIR can surprisingly perform well on the LQ image with unknown and complex degradation, as verified by our experiments. 


\noindent\textbf{Identity semantics} 
To model the identity semantics,  we select a reference face image $Z$ from a candidate image pool by measuring the facial similarity using a face recognition network \cite{deng2018arcface} 
between the candidate and the input LQ face image $X$. 
For the latent space exploration, we adopt the face recognition network $T(\cdot)$ \cite{deng2018arcface} to penalize the facial discrepancy between the estimated HQ face image $\hat{X} = G(w) $ and the reference image $Z$ in the pixel space.
Besides, to improve the efficiency of the exploration,  
we find a guide latent code $w_z \in \mathcal{W}^{+}$ inverted from $Z$ \cite{tov2021designing} to reduce the search space for the latent code $w$. The above process can be formulated by  
\begin{equation}
  \resizebox{.9\hsize}{!}{$\mathcal{L}_{\mathrm{ID}}(Z,w) =  1 - \left\langle T(Z), T(G(w))\right\rangle + \lambda \norm{w-w_z}^2_2$}
  \label{id_loss_1}
\end{equation}
where $\left\langle\cdot,\cdot\right\rangle$ denotes the inner product. 

\noindent\textbf{Emotion semantics}
Following \cite{Guo_2020}, seven facial expressions are used to characterize emotion states including neutral, angry, sadness, happiness, surprise, fear, and disgust.
To find the directions of the different expressions in the latent space, we generate one million face images by randomly sampling latent codes $w$ from the GAN space $\mathcal{W}^{+}$.
By using a facial expression detection tool \cite{msft2021}, we divide the generated faces into seven clusters centered by the predefined seven facial expressions. 
For each cluster, we train a linear binary classifier that determines a boundary hyperplane of the latent representations $w$ by judging the samples in or out of the cluster. 

\begin{figure}[h]
  \centering
  \vspace{-10pt}
  \includegraphics[width=0.5\linewidth]{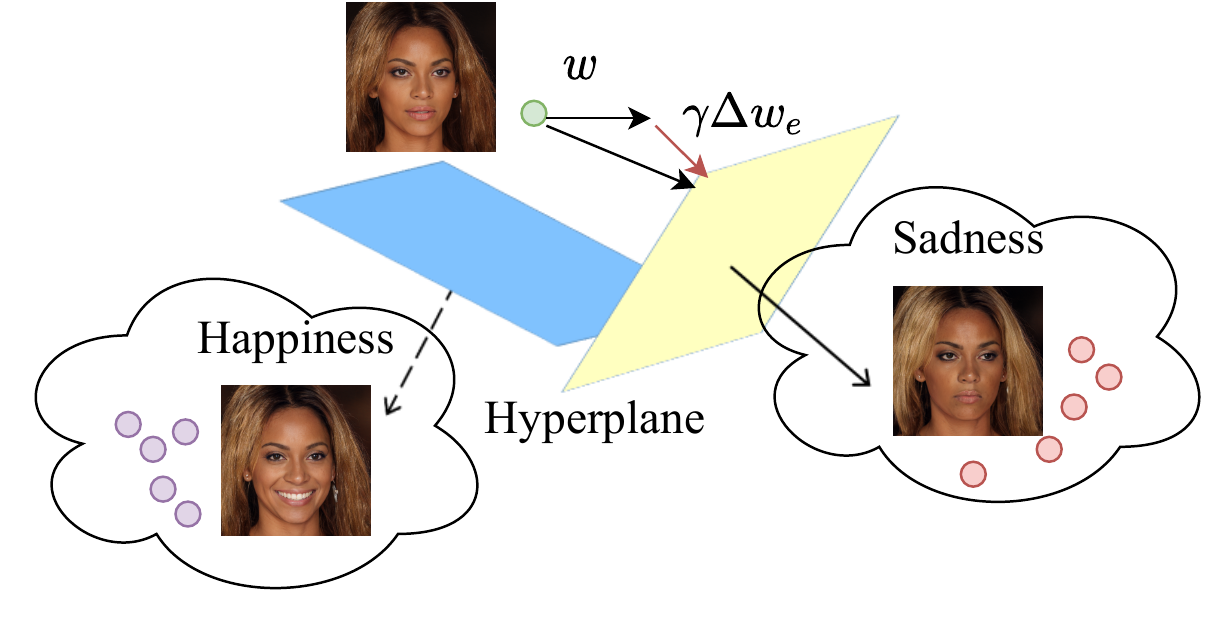}
  \vspace{-10pt}
  \caption{Illustration of the mechanism of the emotion-aware optimization.}
  \label{loss_emotion}
\vspace{-0.5cm}
\end{figure}
In the $\mathcal{W}^{+}$ space, each expression cluster has a direction vector $\Delta w_e$ towards the cluster center. 
As depicted in Fig.~\ref{loss_emotion}, the normal vector of this boundary hyperplane is the direction vector $\Delta w_e$ that guides the optimization along the emotion's direction. Denoting the optimization step size by $\gamma$, the objective to achieve the expected expression is defined as follows
\vspace{-5pt}
\begin{equation}
  \mathcal{L}_e(w + \gamma \Delta w_e) = 1 - \frac{{(w + \gamma \Delta w_e)}^T\cdot\Delta w_e}{\norm{w + \gamma \Delta w_e}\norm{\Delta w_e}}
  \label{emotion_loss}
\end{equation}
\noindent\textbf{Skin color semantics}
To match the facial skin color, 
we adopt a histogram loss\footnote{We adopt a differentiable histogram calculation \cite{DBLP2021} that approximates the hard-binning. } $\mathcal{L}_{hist}$ to penalize
the skin color deviation between the LQ face image $Y$ and the estimated HQ face image. 
To factor out the background influence, before calculating the histogram, we segment the face part of the generated face image $G(w)$ by the detected face segmentation mask $\mathcal{F}_G$. 
The objective function $\mathcal{L}_{hist}$ is defined as follows 
\vspace{-5pt}
\begin{equation}
  \begin{aligned}
    \resizebox{.9\hsize}{!}{$\mathcal{L}_{hist}(G(w),Y) =  \mathlarger{\int} |  CDF(\mathcal{F}_G\odot G(w)) - CDF(\mathcal{F}_G\odot Y)|$}
  \end{aligned}
\vspace{-5pt}
\end{equation}
where $\odot$ represents the Hadamard product. $CDF(\cdot)$ represents the cumulative distribution function of the histogram.

In summary, by rewriting Eq.~(\ref{eq_minmum_w}), we minimize the total loss $f(w)$ which is formulated by 
\begin{equation}
  \begin{aligned}
  \min_{w\in\mathcal{W}^{+}} \mathcal{L}_{d}(Y,w)  + \underbrace{\mathcal{L}_{\rm{ID}} + \alpha_1 \mathcal{L}_e + \alpha_2 \mathcal{L}_{hist} }_{\mathcal{L}_{\mathrm{SI}}(\mathrm{SI}(R,Y),\mathrm{SI}(G(w)))}
  \end{aligned}
  \label{eq_minmum_fw}
  \vspace{-5pt}
\end{equation}
As all the components in Eq.~(\ref{eq_minmum_fw}) are differentiable, we can solve this optimization problem through gradient descent, by back-propagating the gradient of the objective function. 

\begin{figure*}[htbp]
  \centering
  \includegraphics[width=0.9\linewidth]{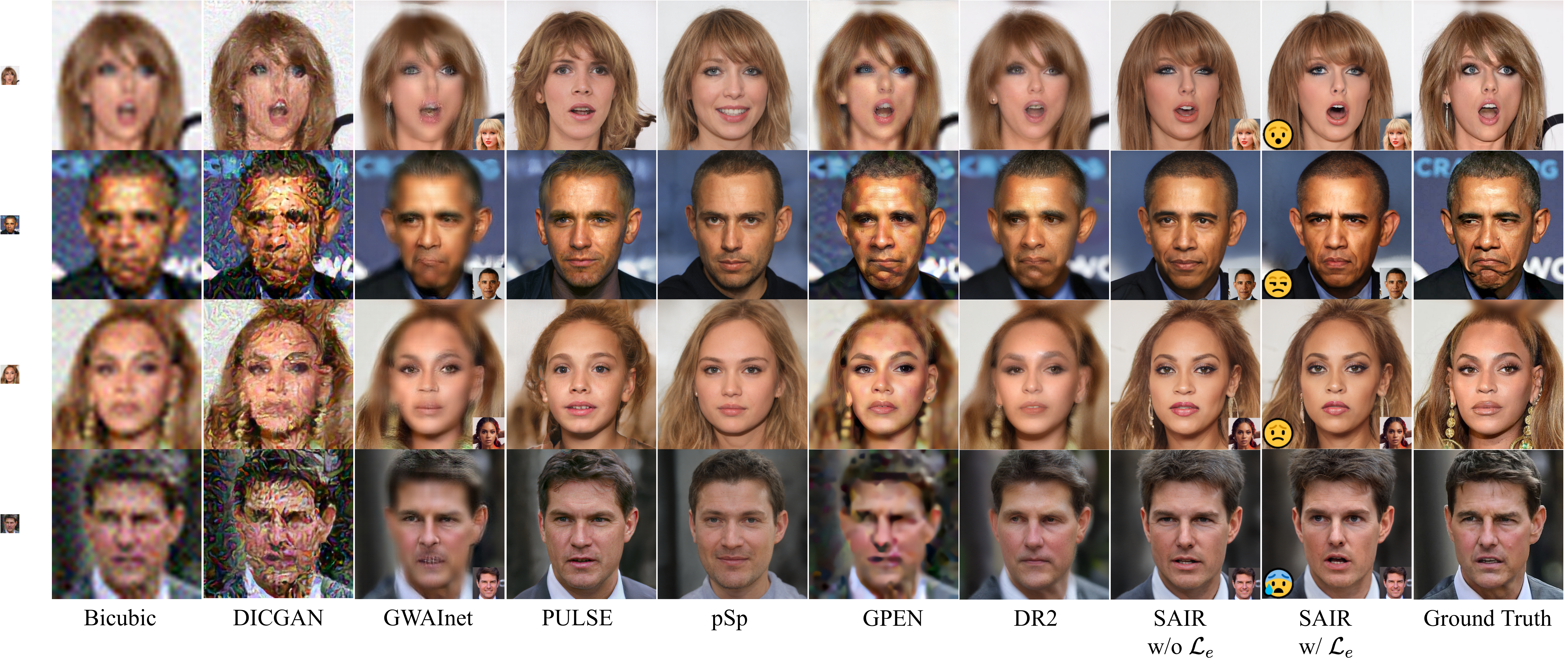}
  \vspace{-0.4cm}
  \caption{Qualitative comparison between our proposed SAIR and the state-of-the-art methods on the LQ face images with unknown noise. The input LQ face images ($32\times 32$) are in the first column.}
 \vspace{-0.5cm}
  \label{fig_comparison}
\end{figure*}

\section{Experiments}\label{exp_label}
\vspace{-0.1cm}
In this section, we quantitatively and qualitatively compare SAIR  with state-of-the-art competitors for face image superresolution. 
In addition, we evaluate the robustness of SAIR. 
To build the test set, we collect 400 HQ face images (resized to $1024 \times 1024$) of celebrities with different expressions and poses from Internet and for each celebrity we randomly choose one face image to build the reference image pool. For optimization, 
we use the Adam optimizer with a learning rate of 0.1, $\beta_1=0.9$, $\beta_2= 0.999$. 
We empirically set the parameters $\lambda=0.001$,  $\alpha_1 = 0.1$ and $\alpha_2=0.05$.
To speed up the convergency, we set the initial $w=w_z$. We set the total iteration number to 400.
To restore a $1024\times 1024$ HQ face image, SAIR needs about 100 seconds on a single GPU (TITAN Xp).

\vspace{2pt}
\noindent\textbf{Face Super-Resolution}
We downsample the test images with $\times 32$ bicubic downsampling followed by a zero-mean gaussian noise with random variance from 0.001 to 0.003. 
SAIR and the competitors that include DICGAN \cite{ma2020deep}, GWAInet \cite{dogan2019exemplar}, PULSE \cite{menon2020pulse}, pSp \cite{richardson2020encoding}, GPEN \cite{Yang2021GPEN}, and DR2 \cite{wang2023dr2}\footnote{Both DICGAN and GWAInet are of CNN network trained by the GAN loss. The GWAInet accepts an extra guided face image to help improve its performance. PULSE and pSp are GAN prior based methods. DR2 is a Diffusion prior based method.}
are required to blindly restore HQ face images ($\times 32$ upsampling) from the LQ images with unknown random noise. 
We directly use their pre-trained models for the comparison. 

As Fig.~\ref{fig_comparison} shows, DICGAN and GWAInet fail to restore clean results. Even worse, DICGAN produces weird artifacts in the results, as in most end-to-end trained CNN models that fail to remove unseen noise even if the noise level is not high. 
In contrast, the GAN prior based methods PULSE and pSp can produce visually good results. But they confuse semantics, deviating too much from the true identity of the input LQ faces. The GPEN and DR2 generate results faithful to the input but blurred. 
Only the proposed SAIR can restore the highest quality faces while preserving the
identity of the LQ images. Furthermore, SAIR produces more correct expressions with emotion-aware semantics \eg the surprise expression of the 1st-row result and the disgust expression of the 2nd-row result. 

\begin{figure}[H]
\vspace{-0.3cm}
  \centering
  \scalebox{0.6}{
  \begin{tabular}{|c|c|c|c|} \hline
    Method & FID $\downarrow$  & LPIPS $\downarrow$ & Similarity $\uparrow$ \\ \hline\hline
    Bicubic      & 252.1 & 0.544  & 0.47 \\
    DICGAN (CVPR 2020 \cite{ma2020deep}) & 329.7   & 0.598  &  0.35  \\
    GWAInet (CVPRW 2019 \cite{dogan2019exemplar})     &  168.8   &0.420    & 0.48 \\
    Pulse (CVPR 2020 \cite{menon2020pulse})       &  143.1  &0.382     & 0.16 \\
    pSp (CVPR 2021 \cite{richardson2020encoding})     & 151.6 &0.390 & 0.11\\
    GPEN (CVPR 2021 \cite{Yang2021GPEN}) & 171.9 &0.403 & 0.27 \\
    DR2 (CVPR 2023 \cite{wang2023dr2}) & 155.1 &0.415 & 0.48 \\
    SAIR     & \textbf{108.7}&  \textbf{0.373} & \textbf{0.54}\\
  \hline
  \end{tabular}  
  }  
\vspace{-0.2cm}
\captionof{table}{Quantitative comparison of different face super-resolution methods.}
\label{tab_comparison}
\vspace{-0.5cm}
\end{figure}

For the quantitative comparison, we adopt the Fr\'{e}chet Inception Distance (FID) \cite{heusel2018gans}, the LPIPS \cite{zhang2018perceptual}
score and the face similarity score \cite{deng2018arcface}
between the result and the ground truth. 
Table.\ref{tab_comparison} reports the quantitative results. 
The lower FID score and LPIPS score mean better perceptual performance. The higher similarity score means more correct identity.

\begin{figure}[H]
\vspace{-10pt}
\begin{minipage}{\textwidth}
\begin{minipage}[b]{0.5\linewidth}
\centering
  \includegraphics[width=0.7\textwidth]{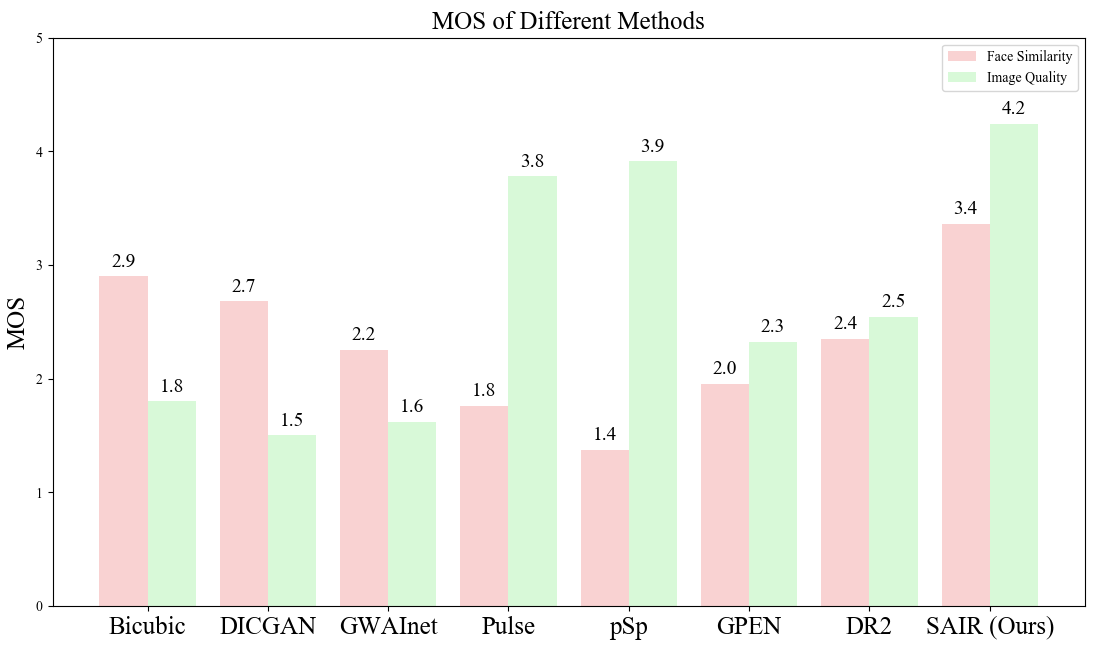}
  \vspace{-10pt}
  \captionof{figure}{User study of different image restoration methods.}
  \label{userstudy}
\end{minipage}
\end{minipage}
\end{figure}

\noindent\textbf{User Study} 
We conduct a user study as a subjective evaluation of our method and the competitors. We adopt the mean-opinion-score (MOS) \cite{ledig2017photorealistic} test. 
For evaluation, we randomly choose 50 images from the collected test set and downsample them 
to $32\times32$ LQ images. 
All the results of the methods in Table.\ref{tab_comparison} and the corresponding ground truth images are presented in a random order to 15 volunteers\footnote{Volunteers are asked to mark the face similarity and the image quality in the range of 1$\sim$5, where  
rating 1 (worst) is exemplified by nearest-neighbors upsampling, and rating 5 (best) is exemplified by the ground truth images.} for marking. 
Finally, we obtained 750 valid evaluation samples. 
As Fig.~\ref{userstudy} shows, SAIR outperforms all the competitors in both the face similarity and quality scores. 

\begin{figure}[h]
\vspace{-20pt}
  \centering
  \includegraphics[width=0.9\linewidth]{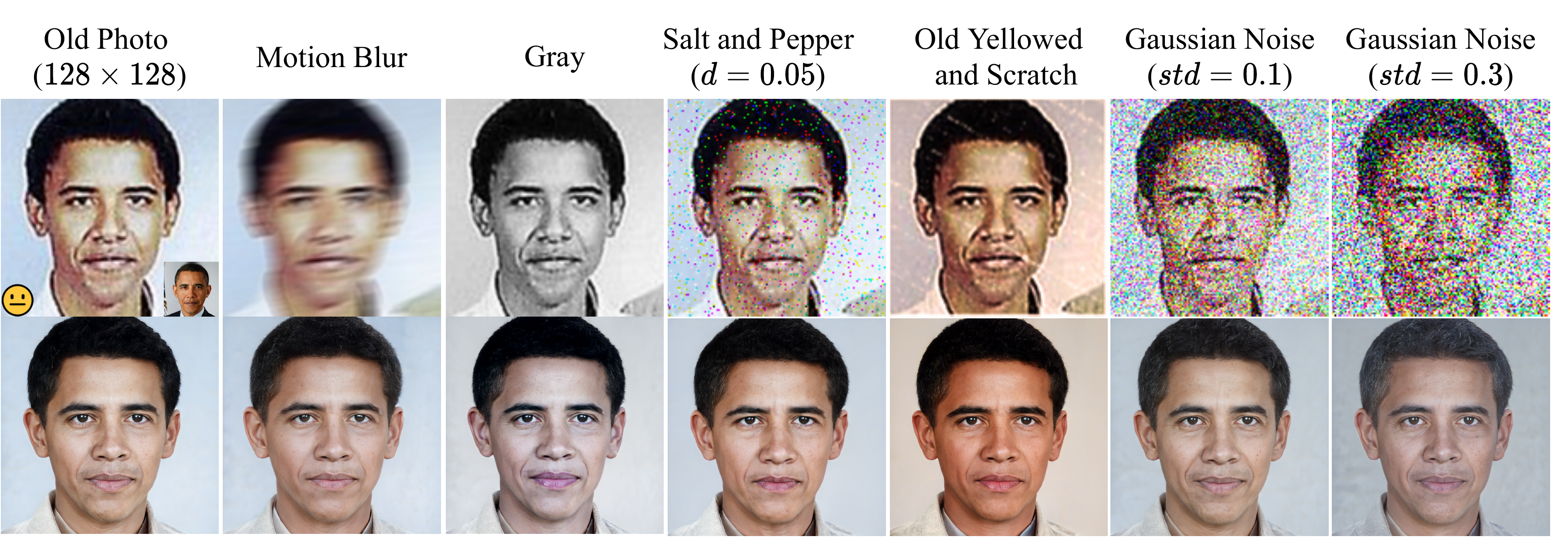}
  \vspace{-10pt}
  \caption[short]{Illustration of the robustness evaluation with various degradation models.}
  \label{robust_eval}
  \vspace{-10pt}
\end{figure}

\begin{figure}
\vspace{-10pt}
  \centering
  \includegraphics[width=0.8\linewidth]{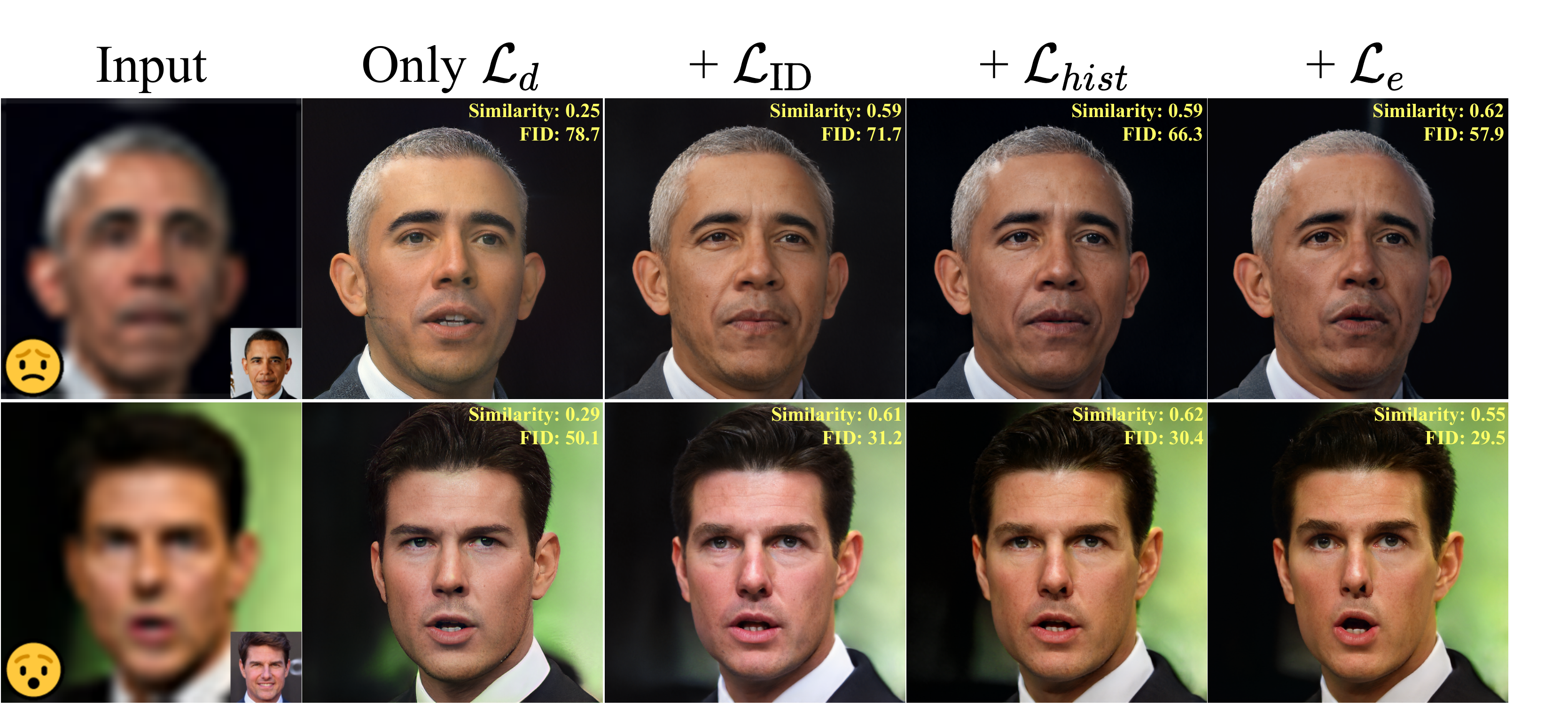}
  \caption{From left to right, the results are produced by gradually adding the priors. The rightmost column is the full version.}
\label{fig_ablation}
\vspace{-15pt}
\end{figure}

\noindent\textbf{Evaluation of the Robustness}\label{robust_sec}
Since SAIR needs a differentiable degradation model $D(\cdot)$, there is a question of whether SAIR works well when encountering unknown degradation models. 
To evaluate the robustness of SAIR, we randomly select 100 test face images and generate the LQ faces by downsampling them to $128\times 128$ resolution followed by various degradation models. 
Then we restore the clean HQ images by $\times 8$ upsampling using SAIR. 
As shown in Fig.~\ref{robust_eval}, even though, the set degradation model $D(\cdot)$ of SAIR is the simple bicubic downsampling followed by a JPEG compression, 
SAIR can impressively maintain good consistency in restoring HQ face images from the various LQ face images, highlighting itself by robust image restoration ability. 

\noindent\textbf{Ablation Study}
We conduct ablation studies on each prior used in SAIR. As shown in Fig.~\ref{fig_ablation}, the column of ``only $\mathcal{L}_d$" presents that the recovered face images have wrong identities as it confuses 
identity semantics. 
By adding $\mathcal{L}_{\rm{ID}}$, the restored face images can possess more correct identities.  
The $\mathcal{L}_{hist}$ makes the skin color of the recovered HQ faces closer to the true face. The rightest
column in Fig.~\ref{fig_ablation} reveals the effect of $\mathcal{L}_e$ making the expression of the estimated HQ faces closer to the intended facial expressions inferred from the LQ images.

\section{Conclusion}

In this paper, we propose SAIR, a self-supervised latent space exploration method for face image restoration, by explicitly modeling the side semantics from a one-shot reference. 
Qualitative and Quantitative experiments demonstrate that SAIR stands out from prior arts in its ability to restore perceptually superior face images with correct semantics. SAIR is scalable and able to be applied to LQ images with complex unknown degradations. 
Albeit focusing on face images in this work, we can easily extend SAIR to recover other types of images by adopting different latent spaces and semantic references.

\bibliographystyle{IEEEbib}
\bibliography{main}

\end{document}